\documentclass[runningheads]{llncs}
\usepackage[T1]{fontenc}
\usepackage{amsmath} 
\usepackage{amssymb} 
\usepackage{booktabs}
\usepackage{hyperref}
\usepackage{multirow}
\usepackage{xcolor}
\usepackage{graphicx,verbatim}
\usepackage{color}

\begin{document}
\title{NimbleReg: A light-weight deep-learning framework for diffeomorphic image registration}
\titlerunning{NimbleReg: Lightweight diffeomorphic image registration with deep-learning}
\author{Antoine Legouhy\inst{1} \and  Ross Callaghan\inst{2} 
 \and Nolah Mazet\inst{3} \and Vivien Julienne\inst{3} \and
Hojjat Azadbakht\inst{2} \and Hui Zhang\inst{1}}
\authorrunning{A. Legouhy et al.}
\institute{Hawkes Institute \& Department of Computer Science, University College London, London, UK 
\and AINOSTICS ltd., Manchester, UK
\and ESIR, Universit\'e de Rennes, Rennes, France}
    
\maketitle              % typeset the header of the contribution
\begin{abstract}
This paper presents NimbleReg, a light-weight deep-learning (DL) framework for diffeomorphic image registration leveraging surface representation of multiple segmented anatomical regions. Deep learning has revolutionized image registration but most methods typically rely on cumbersome gridded representations, leading to hardware-intensive models. Reliable fine-grained segmentations, that are now accessible at low cost, are often used to guide the alignment.
Light-weight methods representing segmentations in terms of boundary surfaces have been proposed, but they lack mechanism to support the fusion of multiple regional mappings into an overall diffeomorphic transformation. Building on these advances, we propose a DL registration method capable of aligning surfaces from multiple segmented regions to generate an overall diffeomorphic transformation for the whole ambient space. The proposed model is light-weight thanks to a PointNet backbone. Diffeomoprhic properties are guaranteed by taking advantage of the stationary velocity field parametrization of diffeomorphisms. We demonstrate that this approach achieves alignment comparable to state-of-the-art DL-based registration techniques that consume images.
\end{abstract}

\section{Introduction}
Non-linear image registration is a fundamental tool in medical image analysis. It is of particular importance in computational neuroanatomy where we seek highly non-linear yet diffeomorphic mappings to produce accurate anatomical correspondences between brain scans of different individuals or between an individual brain scan with a standard reference atlas. Although image registration methods based on conventional optimization algorithms have achieved wide uptake, due to their iterative nature, they generally require substantial computational time.  To address this, alternative implementations exploiting deep learning (DL) have been developed.  DL-based methods enable one-shot, near-instantaneous estimation of anatomical correspondences, with approaches based on unsupervised (a.k.a. self-supervised) training achieving the most success~\cite{devos2017,devos2019,hu2018}.  However, one limitation of DL-based methods is that they are not light-weight, requiring large amount of memory, because they represent input images, intermediate feature maps, and target deformation fields all in terms of dense 3D grids.

The aim of this work is to develop a light-weight DL-based image registration algorithm. Our approach is motivated by the following observations.  First, accurate, and near-instantaneous, image segmentation can now be determined with light-weight DL-based algorithms.  Second, it has been recently shown high-quality image registration can be achieved using a similarity metric computed solely from image segmentation~\cite{hoffmann2021,iglesias2023}.  Third, image segmentation may be efficiently represented in terms of its boundary surface and light-weight DL-based registration of corresponding boundary surfaces now exists~\cite{baum2020,min2024}.

Given the above observations, we propose to recast the image registration problem as one of registering a set of corresponding anatomical structures represented by their respective boundary surfaces. However, existing DL-based algorithms for registering boundary surfaces~\cite{baum2020,min2024} are not equipped with a theoretical framework that supports the fusion of multiple regional mappings into an overall diffeomorphic transformation. To address this, we proposed to leverage the log-Euclidean framework for diffeomorphisms~\cite{arsigny2006} which allows us to efficiently produce and manipulate diffeomorphic transformations through their stationary velocity field (SVF) parametrization.  This framework is not only widely used in traditional~\cite{bossa2007,vercauteren2008} and DL-based~\cite{dalca2019,hoffmann2021,iglesias2023} image registration, but also applied to non-gridded data through kernel convolution~\cite{yang2015}.

The proposed framework, which we call NimbleReg, is light-weight, fast and guarantees diffeomorphic transformations. We evaluate this method on a brain registration task. We compare it against a state-of-the-art DL-based registration technique that processes images. 

\section{Methods}
\label{method}

NimbleReg performs image registration through region-wise alignment of anatomical structures represented by their respective boundary surfaces. Such surfaces can be obtained from a segmentation following the preprocessing steps in Sec~\ref{preproc}.
It consists of two main components. First, a trainable registration model $f_\theta$ takes as input two point clouds associated with the moving and reference surfaces of a given anatomical structure, and outputs a set of velocity vectors associated with the moving point cloud (see Sec~\ref{veloc}). 
Second, an integrator constructs a stationary velocity field (SVF) from the sparse velocities produced by $f_\theta$ through convolution, which is then integrated to generate a diffeomorphic transformation for the whole ambient space (see Sec~\ref{integ}).

\begin{figure}
    \centering
    \includegraphics[width=\linewidth]{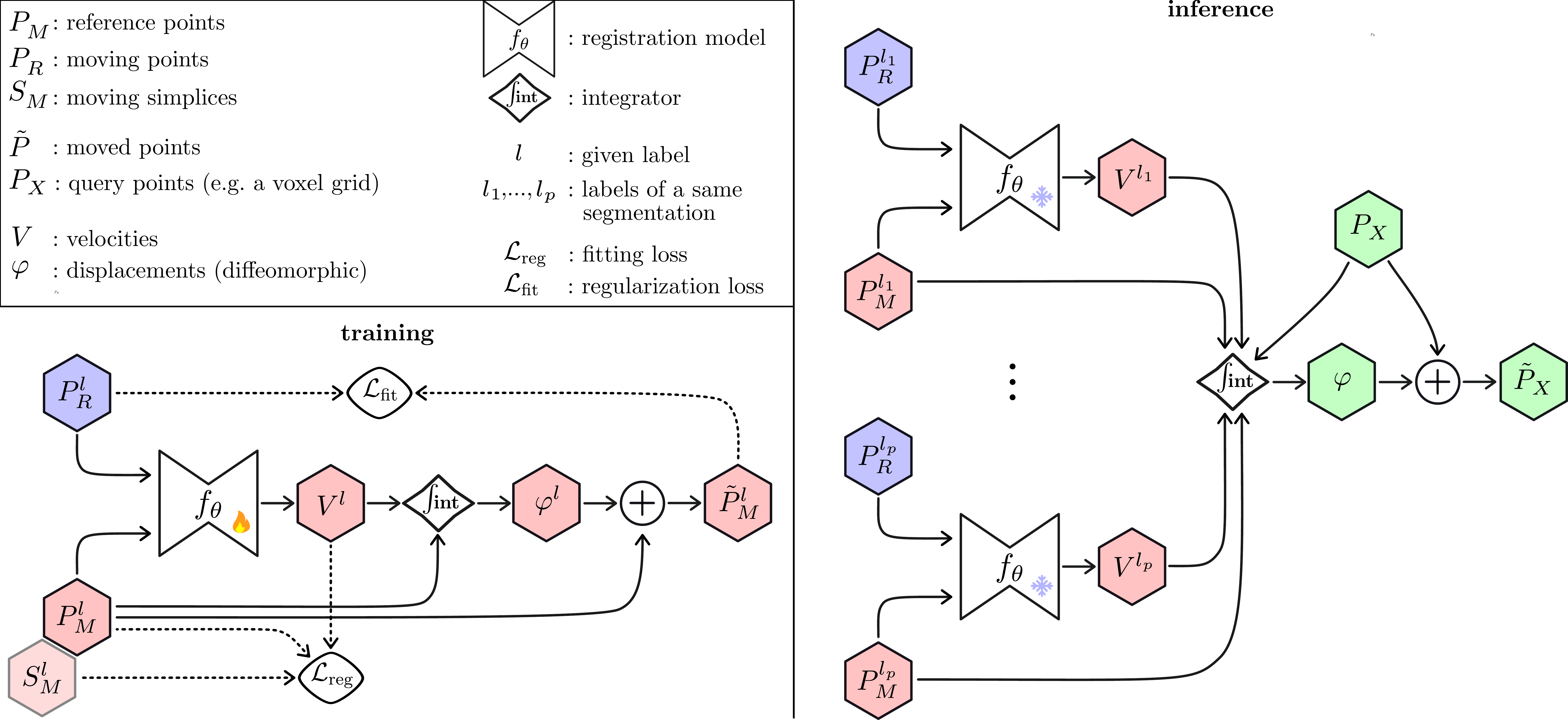}
    \caption{Diagrams for the proposed registration model at training (bottom-left) and at inference (right), and a zoom in on $f_\theta$'s architecture (top-left).}
    \label{model_diag}
\end{figure}

\paragraph{Training:} Training is performed in a self-supervised manner. For a given region, the moving and reference surface points are fed to $f_\theta$, which outputs velocity vectors for the moving points. The resulting velocity vectors are used by the integrator to produce displacement vectors for the moving points. The quality of the alignment between the moved and reference points is assessed with a fitting loss. A regularization loss is computed from the moved points and associated simplex structure (Eq.~\ref{regul}). See Fig.~\ref{model_diag}, bottom-left panel.

\begin{figure}
    \centering
    \includegraphics[width=\linewidth]{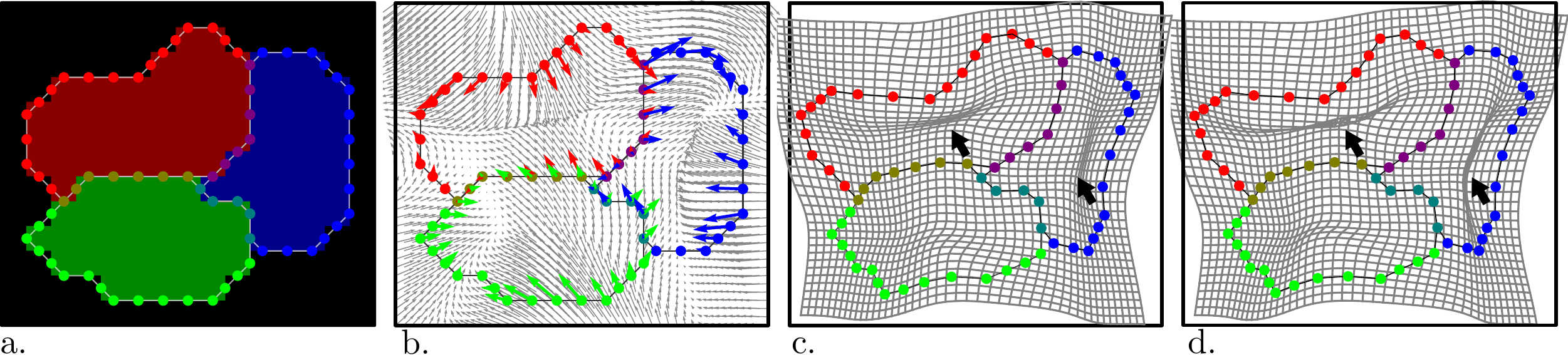}
    \caption{a) Synthetic segmentation and associated extracted surfaces. b) Velocities estimated by $f_\theta$ for each region and the associated SVF; c) Deformed points and grid after integration. d) Deformed points and grid without integration.}
    \label{diag}
\end{figure}

\paragraph{Inference:} Moving and reference points for each considered region are fed in turn to $f_\theta$ to produce velocities for each region (see Fig.~\ref{diag}.b). The moving points and velocities for all regions are used by the integrator to estimate displacements for any query points (e.g. the points of an image grid) (see Fig.~\ref{diag}.d-e). The SVF parametrization of diffeomorphisms~\cite{arsigny2006} allows us to seamlessly blend the velocities of all regions to produce a diffeomorphic transformation (see Fig.~\ref{model_diag}, right panel). Because the inverse of the transformation exists and can be computed easily, one can apply the transformation to an image (backward coordinate mapping). 

\subsection{Preprocessing}
\label{preproc}
Given an image segmented into multiple anatomical regions, the following preprocessing steps are used to extract region surfaces and perform an initial alignment to a standard reference.
\begin{enumerate}
    \item A surface for each labeled region is extracted using a tool like marching cubes~\cite{lorensen1987}, with an implementation ensuring that shared interfaces between adjacent regions produce identical points on their respective surfaces (see Fig.~\ref{diag}.a).
    \item The individual region surfaces are then merged into a single mesh, stitched together using the common interface points.
    \item The overall mesh is smoothed. This approach, rather than smoothing the individual surfaces independently, ensures that the interfaces between adjacent regions are preserved.
    \item A pre-alignment onto a standard reference is performed using Polaffini~\cite{polaffini}, which estimates an initial polyaffine transformation~\cite{arsigny2009} based on the centroids of the labeled regions. This pre-alignment reduces spatial variability in the input domain of the registration model, simplifying the learning process by eliminating the need to account for pose or low-frequency shifts.
    \item The mesh is re-split into individual anatomical surfaces.
\end{enumerate}
Each obtained surface is composed of a tuple of points (point cloud) and a tuple of simplices. The subsequent model takes as input moving and reference surface point clouds from the same region. The simplices of the moving surface are used in the regularization loss.

\subsection{Velocities estimation}
\label{veloc}

A trainable model $f_\theta$ consumes a moving point cloud $p$ and a reference one $q$ associated with surfaces of a given anatomical region; and outputs velocity vectors $v$ for the moving points. The same model is used to process all the regions independently. 
% A model $f_\theta$ with trainable parameters $\theta$ to consume a moving point cloud $p=\left(p_i,\ p_i\in \mathbb{R}^d\right)_{i=1}^n$ and a reference one $q=\left(q_i,\ q_i\in \mathbb{R}^d\right)_{i=1}^m$; and output velocity vectors for the moving points: $v=\left(v_i,\ v_i\in \mathbb{R}^d\right)_{i=1}^n$. 

Our model uses a PointNet~\cite{qi2017} backbone, which is designed to processes point clouds by sliding a shared multilayer perceptron (MLP) over each point to learn local features independently. These features are then aggregated through max pooling to capture a global feature representation of the entire cloud in a point permutation invariant fashion. 
Similarly to~\cite{baum2020,min2024}, $p$ and $q$ are each fed to a common (shared weights) PointNet feature extraction module. The two legs then rejoin by concatenating the global features of both to local features of the moving points. Another round of sliding shared MLP then produce the output velocity vectors attached to each moving point. 

In~\cite{baum2020,min2024}, $v$ is directly used to deform the moving points, which does not provide options for seamlessly fusing multiple regions while guaranteeing diffeomorphic properties. Here, instead, $v$ is treated as a tuple of velocity vectors that parametrizes a velocity field.

\subsection{Integration into a diffeomorphic transformation}
\label{integ}
The goal of the integration module is two-fold: to estimate a smooth velocity field $V$ from sparse set of velocities ouput from $f_\theta$, and to integrate $V$ to obtain a diffeomorphic transformation $\varphi$ for the whole ambient space. This represents a key novelty of our approach, distinguishing it from the work of~\cite{baum2020,min2024}.
% $K_{\sigma}(x, p_i) = e^{-\frac{\|x - p_i\|^2}{2 \sigma^2}}$

Given the sparse sample of velocities $v$ attached to the control points $p$, one can construct a smooth stationary velocity field (SVF) $V$ through kernel convolution~\cite{garcia2010,yang2015}. Let $K_\sigma$ a kernel of bandwidth $\sigma$ that smoothly decays with distance. $V$ can be evaluated for all $x$ of the ambient space according to: $V(x)=\frac{\sum _{i}K_{\sigma}(\|x-p_{i}\|)v_{i}}{\varepsilon+\sum _{i}K_{\sigma}(\|x-p_{i}\|)}$, with $\sigma$ controlling the smoothness.

The log-Euclidean framework~\cite{arsigny2006} for diffeomorphisms confers an infinite Lie group structure to diffeomorphisms parametrized by stationary velocity fields (SVF). A sufficiently smooth stationary velocity field $V$ defines a one-parameter subgroup of diffeomorphisms $\varphi^t$, given as the unique solution of the stationary ordinary differential equation: $\frac{d\varphi^t}{dt}=V(\varphi^t)$ with initial condition $\varphi^0=\mathrm{Id}$. Integrating this equation during one unit of time defines the Lie exponential map, which yields the final transformation: $\exp(V)=\varphi^1$, simply noted $\varphi$. Through this Lie group embedding, one can perform linear operations on the Lie algebra and then map the result into a diffeomorphism using the exponential. The inverse transformation can be computed easily through: $\varphi^{-1}=\exp(-V)$.
The integration can be performed using numerical analysis methods like Runge-Kutta.

This integrator takes as input a set of control points $p$ with their associated velocities $v$, along with a set of query points $x$ where the estimation is to be performed. It outputs a displacement for each point in $x$, corresponding to the diffeomorphic transformation $\varphi$. Multiple outputs from $f_\theta$, and corresponding multiple anatomical regions, can be used as input. In this context, points at the interface between adjacent regions will be associated with more than one velocity vector. Yet, the integration of the resulting SVF from convolution will produce a diffeomorphic transformation.

Since Polaffini~\cite{polaffini} also utilises the SVF framework, $\varphi$ and the initial transformation can be easily manipulated together and ensure an overall diffeomorphic transformation. Given $\psi_p$ and $\psi_q$, the transformations from the Polaffini pre-alignment for the moving and reference respectively, the full transformation $T$ going from moving to reference can be retrieved as: $T=\psi_p\circ \varphi \circ \psi_q^{-1}$. One can even take advantage of the Baker-Campbell-Hausdorff approximation~\cite{bossa2007,vercauteren2008}.

\subsection{Loss functions}
To define a distance metric between point clouds without one-to-one correspondences, a common heuristic from the Iterative Closest Point (ICP) algorithm~\cite{besl1992} is to use nearest-neighbor projection to establish correspondences. Similarly to~\cite{baum2020,min2024}, we adopt the symmetric formulation, known as Chamfer distance, as fitting loss $\mathcal{L}_{\text{fit}}$ to assess the quality of alignment between the transformed and reference point clouds.
% \begin{equation}
% \label{chamfer}
% \mathcal{L}_{\text{fit}}(p,q) = \frac{1}{2|p|} \sum_{p_i\in p} \min_{q_j\in q} \|p_i-q_j\|^2
% + \frac{1}{2|q|} \sum_{q_j\in q} \min_{p_i\in p} \|p_i-q_j\|^2
% \end{equation}

We adopt a regularization loss that differs from~\cite{baum2020,min2024}. It is designed to promote smooth velocities $v$ along the surface of the considered structure by leveraging the simplex information of the surface as auxiliary data. Given the tuple of simplices $s$ associated with the moving surface, abrupt changes of velocity between points of the same simplex are penalized following:

\begin{equation}
\label{regul}
\mathcal{L}_{\text{reg}}(p,s,v) = \frac{1}{|s|} \sum_{k=1}^{|s|} \sum_{\substack{i,j \in s_k\\ i \neq j}} \frac{\| v_i - v_j \|^2}{\| x_i - x_j \|^2}
\end{equation}

The overall loss is: $\mathcal{L}=\mathcal{L}_{\text{fit}} + \lambda\mathcal{L}_{\text{reg}}$ where $\lambda$ weights the regularization.

\section{Experiments and results}
\subsection{Evaluation strategies}
Our proposed method, NimbleReg, is evaluated at a non-linear brain image registration task against EasyReg~\cite{iglesias2023}, a state-of-the-art registration tool based on SynthMorph~\cite{hoffmann2021}. 
We also included in the comparison two initializations based on region centroids using Polaffini~\cite{polaffini}: Affine ($\sigma=\infty$, similar to~\cite{iglesias2023}) and Polyaffine ($\sigma$ following Silverman's rule of thumb).

We evaluated the quality of the alignment using two metrics. The first metric is the Jaccard index (IoU), a volume-based metric that measures the overlap between the moved and reference segmentations. The second metric is the Chamfer distance, a surface-based metric, between moved and reference interfaces. From the preprocessing steps in~\ref{preproc}, one can extract surface interfaces between regions, which is more fine-grained than whole region surfaces. This distance is computed for each pair of regions (including the background) that are adjacent both in the moving and reference images. We also compared the computational burden of EasyReg and NimbleReg in terms of memory usage.

\subsection{Data}
We used T1-weighted brain MRI data from three datasets to cover various brain maturation and health conditions: (training/validation/testing): ADNI~\cite{adni} (\url{adni.loni.usc.edu}), UK Biobank (\url{ukbiobank.ac.uk}) and IXI~(\url{brain-development.org/ixi-dataset}).
Subjects were divided into training/validation/testing sets of following sizes: ADNI: 60/15/150 (equal number of AD, MCI and HC), UK Biobank: 20/5/100 and IXI: 20/5/100. All images are 3D and roughly have 1 mm isotropic voxel size.
The images were segmented according to the DKT~\cite{dkt1,dkt2} protocol. SynthSeg~\cite{synthseg} was used to match the EasyReg pipeline, but light-weight alternatives, e.g. FastSurfer~\cite{henschel2020}, achieve similar results for T1-weighted images. Left and right cerebral white matter, and outer CSF were ignored, leaving 95 regions to register.
During training, random pairs of moving and reference corresponding anatomical region surfaces were selected from their respective datasets for both training and validation.
Each subject in the test set was used as the moving image once, while the reference image was randomly selected from a permutation of the same set. This led to 350 unique pairs to be registered.

\subsection{Implementation details}
\label{implementation}

Surface extraction was performed using marching cubes~\cite{lorensen1987} (vtkDiscreteMarchingCubes\footnote{\label{note1}The Visualization Toolkit (VTK), open source: \url{vtk.org}.}), with simplices being triangular faces. Smoothing was done using a windowed sinc function interpolation kernel (vtkWindowedSincPolyDataFilter\footref{note1} with 100 iterations). Pre-alignment was performed onto an MNI template~\cite{fonov2009} using Polaffini~\cite{polaffini} with $\sigma$ following Silverman's rule of thumb. We used decimation (vtkQuadricDecimation\footref{note1}) and random duplication to obtain fixed numbers of 1100 points and 2000 simplices per region surface. Point coordinates were normalized to $[0,1]$ using the MNI image boundaries.

In the model $f_\theta$, the first sliding MLP to extract local features had layers of sizes (64, 64), the second before global features extraction: (64, 128, 1024), and the third that output the velocities: (1024, 512, 256, 128, 64, 3); ReLU activation, except for the last layer. Adam optimizer was used with learning rate $10^{-4}$. The regularization weight was set to $\lambda=10^{-5}$. The model with the best validation loss after 1000 epochs was selected. We used a batch size of 50.

The standard deviation of the Gaussian kernel in the integrator was set to $\sigma=2.33$ mm ($\sigma=10^{-2}$ in normalized coordinates). Integration was performed using first-order Runge-Kutta (Euler's method) with 12 time steps.

The code is available at \url{github.com/anonimized}.

\subsection{Results}
\begin{figure}
    \label{results}
    \centering
    \includegraphics[width=0.98\linewidth]{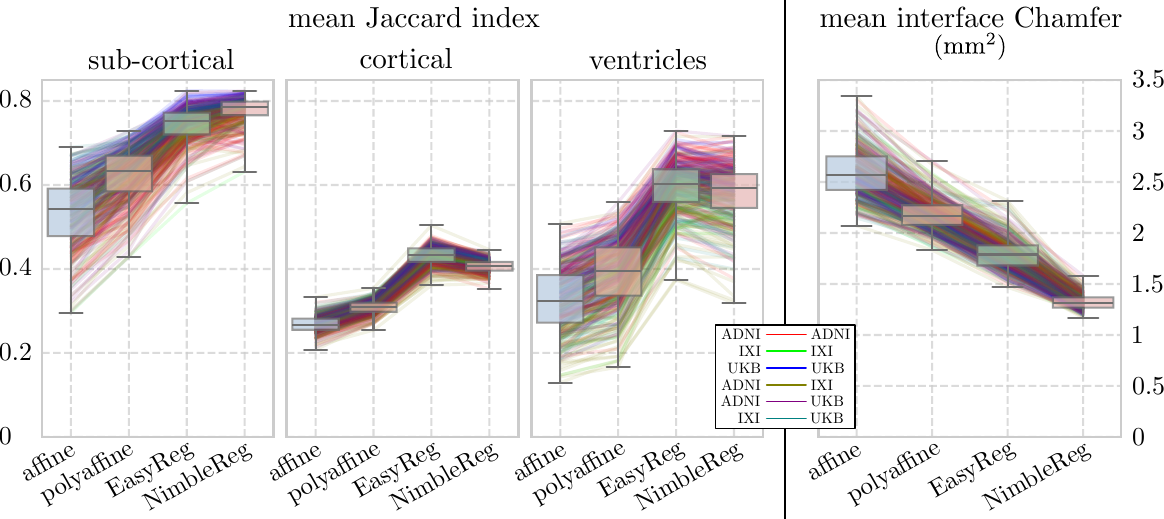}
    \caption{Quality of alignment metrics between moved and reference. Left: segmentation overlap (higher the better). Right: surface distance (lower the better). Lines are colored according to which datasets the images are from.}
\end{figure}

Results for the quality of alignment are shown in Fig.~\ref{results}. For the Jaccard, the scores are averaged over labels, divided into three groups: cortical regions, sub-cortical regions and ventricles. For this metric, we observe comparable performances between EasyReg and NimbleReg for all groups of regions. During training, EasyReg learns to maximize a Dice (close to Jaccard), which is not the case for NimbleReg. The lower accuracy of EasyReg for the surface-based metric should be interpreted with caution due to the asymmetry of the experiment. Since EasyReg is not designed to handle surfaces, the surface extraction was performed after registration, whereas for NimbleReg, the extracted surfaces from the moving image are transformed directly. NimbleReg is trained to minimize the Chamfer distance for regions as a whole, which is close to the interface variant used for evaluation. It is not the case for EasyReg.
The CNN architecture of EasyReg generates heavy feature maps. The ones of the biggest layer weigh about 1.2 GB. At inference, around 24 GB of memory are necessary, only for batch of size 1.
NimbleReg is much more light-weight thanks to its PoinNet backbone. Estimating the velocities through $f_\theta$ is extremely cheap, even 1 GB is more than enough. The biggest memory consumption comes from the integration which necessitates the computation of pairwise distances. However, this burden is alleviated at inference using KD-trees.

\section{Discussion and conclusion}

We presented NimbleReg, a novel deep-learning framework for non-linear image registration. Designed to process region surfaces, it is very light-weight. Using a DL approach based on PointNet to estimate transformation parameters, it enables rapid inference. Leveraging the SVF framework, it seamlessly fuses region-wise deformations into an overall diffeomorphic transformation. The proposed method achieves similar alignment compared to EasyReg, a state-of-the-art DL image registration tool, but with a much lower computational burden. 

The registration model naturally possesses useful symmetries thanks to its PointNet backbone. It is invariant to reference point permutation and equivariant to moving point permutation (the same applies to duplication).
Since a single model is trained to process all regions independently, each region of each subject serves as a sample. As a result, the model is robust to a wide variety of shapes and does not require a large number of training subjects.
Like other segmentation-based approaches~\cite{baum2020,min2024}, cross-modality registration is naturally enabled, provided that the modalities can be segmented. The chosen SVF framework, albeit more rudimentary than its time-varying counterpart~\cite{lddmm}, generally leads to comparable performance in practice~\cite{yang2015}, while being more practical.

Although the results are promising, we believe there is room for improvement in accuracy. Performance in the cortex could benefit from using more points or finer segmentations, as some cortical regions in the DKT atlas are quite large, resulting in severe point decimation.
In addition, the bandwidth of the kernel in the integrator has to be chosen sufficiently large to maintain numerical precision when extrapolating far from control points, which can result in overly smooth transformations hindering alignment. An adaptive $\sigma$ based on point density could improve accuracy.
Finally, we used a rather coarse integration scheme, higher-order Runge-Kutta methods would be beneficial.


\end{document}